\documentclass[acmtog]{acmart}
\acmSubmissionID{727}

\usepackage{float}
\usepackage{xcolor}
\usepackage{fancybox}
\usepackage{makecell}
 \usepackage{multirow}
\usepackage[ruled]{algorithm2e} 

\SetAlFnt{\small}
\SetAlCapFnt{\small}
\SetAlCapNameFnt{\small}
\SetAlCapHSkip{0pt}

\usepackage{graphicx}
\usepackage{tabularx}
\usepackage{amsmath}
\usepackage{booktabs} 

\usepackage{bm}
\usepackage{subfigure}

\newcommand\wsize{0.12}

\setcopyright{rightsretained}

\begin{document}
\title{
AniFaceDrawing: Anime Portrait Exploration during Your Sketching
}

\author{Zhengyu Huang}
\orcid{0000-0002-4279-3008}
\affiliation{%
 \institution{Japan Advanced Institute of Science and Technology}
 \city{Nomi}
 \country{Japan}
}
\email{huang.zhengyu@jaist.ac.jp}
\author{Haoran Xie}
\affiliation{%
 \institution{Japan Advanced Institute of Science and Technology}
 \city{Nomi}
 \country{Japan}
}
\email{xie@jaist.ac.jp}
\author{Tsukasa Fukusato}
\affiliation{%
 \institution{Waseda University}
 \city{Tokyo}
 \country{Japan}
}
\email{tsukasafukusato@waseda.jp}
\author{Kazunori Miyata}
\affiliation{%
 \institution{Japan Advanced Institute of Science and Technology}
 \city{Nomi}
 \country{Japan}
}
\email{miyata@jaist.ac.jp}


\renewcommand\shortauthors{Huang, Z. et al}

\begin{abstract}
In this paper, we focus on how artificial intelligence (AI) can be used to assist users in the creation of anime portraits, that is, converting rough sketches into anime portraits during their sketching process.
The input is a sequence of incomplete freehand sketches that are gradually refined stroke by stroke, while the output is a sequence of high-quality anime portraits that correspond to the input sketches as guidance. 
Although recent GANs can generate high quality images, it is a challenging problem to maintain the high quality of generated images from sketches with a low degree of completion due to ill-posed problems in conditional image generation. 
Even with the latest sketch-to-image (S2I) technology, it is still difficult to create high-quality images from incomplete rough sketches for anime portraits since anime style tend to be more abstract than in realistic style. 
To address this issue, we adopt a latent space exploration of StyleGAN with a two-stage training strategy. 
We consider the input strokes of a freehand sketch to correspond to edge information-related attributes in the latent structural code of StyleGAN, and term the matching between strokes and these attributes ``stroke-level disentanglement.'' 
In the first stage, we trained an image encoder with the pre-trained StyleGAN model as a teacher encoder. In the second stage, we simulated the drawing process of the generated images without any additional data (labels) and trained the sketch encoder for incomplete progressive sketches to generate high-quality portrait images with feature alignment to the disentangled representations in the teacher encoder. We verified the proposed progressive S2I system with both qualitative and quantitative evaluations and achieved high-quality anime portraits from incomplete progressive sketches. 
Our user study proved its effectiveness in art creation assistance for the anime style. 

\end{abstract}


%
%

\maketitle

\newcommand\myhsize{0.13}
\renewcommand\wsize{0.15}
\begin{figure*}
\centering
\includegraphics[width=\linewidth]{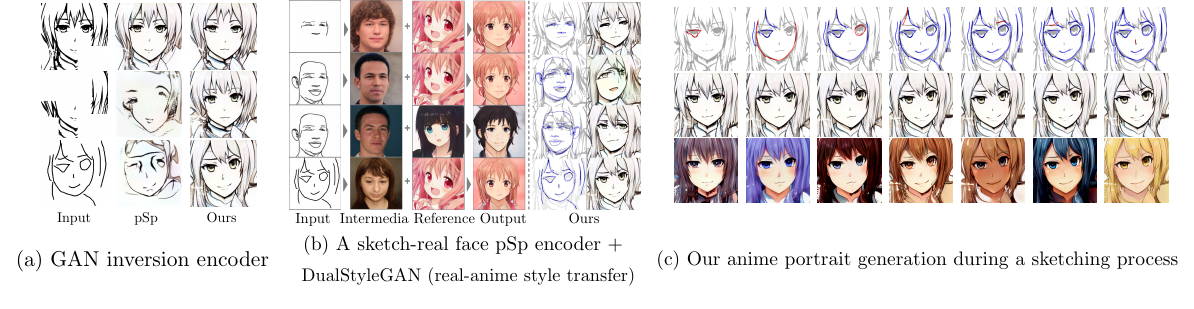}\vspace{-20pt}
\vspace{-2.5mm}
\caption{Comparison of different ideas for sketch-based anime portrait generation. 
In (a), an original pSp encoder, which works for line drawing with small areas missing (first row), cannot correctly recognize user sketches, even for a complete sketch (third row). 
In (b), sketch-anime portrait style transfer using real faces as intermediate results lead to an input/output mismatch and mode collapse. 
And lack of diversity in the results of different inputs fails to assist the user in drawing anime portraits. Realistic sketches (first three rows) are used to obtain better intermediate faces. 
In contrast, our method can generate more diverse anime portraits, even if the input is a realistic sketch after translation (first two rows) or flip (third row). (c) shows that our method can generate high-quality results that consistently match the input sketch throughout the sketching process. To make the matching of sketches and results of our method clear, the intermediate results disentangled most of the color information (second row) are stacked below the input strokes (blue strokes on the first row) once a new stroke (red) is added. 
 The final results (third row) were generated using a random style-mixing technique.  
 Note that all generated results with ``near-white'' hair are intermediate results, which are style mixing with a fixed ``near-white'' color latent code. 
 }
\label{fig:drawingprocess}
\end{figure*}

\section{Introduction}
Using AI to assist the general user in creating a professional anime portrait is not a trivial task. As a popular drawing style,  the anime style is based on realism but has its own characteristics and exaggerations --  line drawings of anime portraits are simpler and more abstract than real human faces. Furthermore,  the input sketches that users make during the drawing process contain little detailed information and lack partial shape information. As a result,  the high-quality synthesis of anime portraits from freehand sketches is challenging. 
The main problem is how to generate appropriate guidance images that match these abstract lines, based on a sequence of incomplete sketches, for the user during freehand sketching. 

We also realized that classic S2I or style transfer techniques would not work for this task. 
Consider a scenario in which a novice tries to create an anime portrait with an ideal AI-assisted system. As the user draws stroke by stroke, the guidance image generated by this system should be able to locally match the sketch as the number of strokes increases. 
However, most S2I approaches tend to consider only complete sketches as input for image generation -- in the case of incomplete sketches, especially those with sparser strokes, they cannot maintain the quality of the output. 
Taking sketch-to-anime-portrait generation with StyleGAN as an example, in \autoref{fig:drawingprocess}(a), the state-of-the-art Pixel2Style2Pixel (pSp)~\cite{psp21} is an encoder for GAN (Generative Adversarial Network) inversion that can successfully reconstruct a complete line drawing into an anime portrait that can tolerate small missing areas (first row), but it gets poor outputs when the input is a sketch with large missing areas (second row) or a rough sketch with less details (third row). 
Therefore, conventional GAN inversion techniques perform poorly in the drawing process -- they do not spontaneously implement stroke-level disentanglement during learning, nor do they naturally maintain partial matches. 
Similarly, the simple combination of a rough sketch-real-face approach with a further style transfer from realistic to anime style may not be a good idea.  \autoref{fig:drawingprocess}(b) 
shows such an example: A sketch-real face pSp encoder is combined with a face-anime style transfer called DualStyleGAN~\shortcite{Yang22DualStyleGAN} for sketch-based anime portrait generation, using real faces as the intermediate results. 
Because the difference between the anime-style face and the real face is relatively large, results generated by this method are not consistent with the input sketches. For the same reason, 
DualStyleGAN itself tends to fall into ``mode collapse,'' 
no matter if the input is a sketch of a realistic style or an anime style.

Thus, we propose a new idea for anime-style portrait generation during sketching. 
Our solution involves sketch-based latent space exploration in a pre-trained StyleGAN~\cite{StyleGAN:Karras19}. 
The advent of StyleGAN made it possible to create high-quality images for many types of subjects, including anime portraits. In turn, this great success led to the rapid development of the GAN control for image editing. 
By applying linear regression to the disentangled latent space of a StyleGAN, users can control the various properties of the generated image by modifying the attribute parameters. In our case, we are trying to implement sketch-based anime portrait generation control during the drawing process, because manipulating multiple shape-related attributes (e.g., pose, mouth shape, or nose position) with separate sliders is not intuitive enough.
Compared with the above-mentioned approaches, our approach allows the generated results to be matched with the users' rough sketches during their drawing process (see \autoref{fig:drawingprocess}(c)). 
To the best of our knowledge, our system is the first to provide anime portraits progressive drawing assistance. 
Our main contributions are summarized as follows:

\begin{itemize}
\setlength{\leftskip}{-3mm}
    \item We present AniFaceDrawing, the first high-quality anime portrait drawing assistance system based on the S2I framework from freehand sketches throughout the entire drawing process. 
    \item We propose an unsupervised stroke-level disentanglement training strategy for StyleGAN, so that rough sketches with sparse strokes can be automatically matched to the corresponding local parts in anime portraits without any semantic labels. 
    \item A user study is conducted to prove the effectiveness and usability of AniFaceDrawing for users when creating anime portraits. 
\end{itemize}

\section{Related work}

\noindent
\textbf{Latent space of StyleGAN}. With the further development of GAN, how to use the latent space to manipulate outputs from pre-trained GANs has also become a hot research topic~\cite{abs-2101-05278}. 
Among various pre-trained GANs, StyleGAN~\cite{sg1_KarrasLA21} is usually the most common choice. 
A typical StyleGAN generator usually involves three types of latent spaces: $\bm{\mathcal{Z}}$, $\bm{\mathcal{W}}$, and $\bm{\mathcal{W}}+$. A random vector $\bm z \in \bm{\mathcal{Z}}$ is often a white noise belonging to a Gaussian distribution, which is the same as the original GAN. 
In StyleGAN, the $\bm z$ vector first passes through a mapping network, which is composed of eight fully-connected layers and is transformed to $\bm w$ embedding into an intermediate latent space $\bm{\mathcal{W}}$. Note that both $\bm z$ and $\bm w$ are 512-dimensional vectors. 
Here, the introduction of this mapping network is to get rid of the influence of the input vector $\bm z$ by the distribution of the input data set and to better disentangle the attributes. Each layer of the StyleGAN generator can receive a vector $\bm w$ of input via AdaIN (adaptive instance normalization). As there are 18 such layers in the StyleGAN generator, StyleGAN can input up to 18 mutually different $\bm w$ vectors. 
This different $\bm w$ can  
be concatenated into a new vector $\bm w+$  with $18 \times 512$ dimensions and the corresponding latent space to $\bm w+$ is called $\bm{\mathcal{W}}+$. 
One application of $w+$ is style mixing, which can also be found in the inference step in \autoref{fig:overview}. 
In addition, we mapped incomplete progressive sketches into the latent space $\bm{\mathcal{W}}+$ of StyleGAN for guidance generation.

\vspace{1.6mm}
\noindent
\textbf{Facial latent space manipulation}. 
One of the most important applications of latent space manipulation is face attribute editing. 
Chiu et al.~\shortcite{ChiuKLIY20} present a human-in-the-loop differential subspace search for exploring the high-dimensional latent space of GAN by letting the user perform searches in 1D subspaces. 
H{\"a}rk{\"o}nen et al.~\shortcite{h2020ganspace} identify latent directions with principal components analysis (PCA), and created interpretable controls for image synthesis, such as viewpoint changing, lighting, and aging. By determining facial semantic boundaries with a trained linear SVM (support vector machine), Shen et al.~\shortcite{InterFaceGAN22} is able to control the expression and pose of faces. 
An instance-aware latent-space search (IALS) is performed to find semantic directions for disentangled attribute editing~\cite{han2021IALS}. 
Instead of the tedious 1D adjustment of each face attribute, we directly use progressive rough sketches to control the shape attributes of the face and explore latent space. 

\vspace{1.6mm}
\noindent
\textbf{GAN-control with encoder-based manipulation}. The pSp encoder implements GAN inversion without optimization by using feature pyramids and mapping networks. Since this method does not need to compute losses between inputs and outputs of a GAN, it can also handle semantic layouts or line drawings as input. To improve the editability of the encoder-based approach, Tov et al.~\shortcite{TovANPC21} introduced regularization and adversarial losses for latent codes into encoder training. In addition, the ReStyle encoder~\cite{ReStyle21} has improved the reconstruction quality of inverted images by iteratively refining latent codes from the encoder. Unlike encoder-based approaches, which require many training pairs, our method automatically generates sketches directly from the GAN on the fly in the training step without additional pairwise data generation.

\vspace{1.6mm}
\noindent
\textbf{Interactive AI assistance}. 
With the rapid development of deep learning, many efforts have been made to apply AI to interactively assist users in various fields, such as music creation~\cite{MusicCreation:Frid20}, handwritten text editing~\cite{Aksan18DeepWriting_s}, and sketch colorization~\cite{Ren20Colorization}. 
When it comes to sketch-based drawing assistance, the dominant idea has remained to adopt retrieval-based approaches~\cite{choi2019sketchhelper,Collomosse19LiveSketch} since the ShadowDraw~\shortcite{lee2011shadowdraw} have been proposed. For example, to improve users’ final sketches, DeepFaceDrawing~\shortcite{DeepFaceDrawingChenS0XF20} and DrawingInStyles~\shortcite{DrawingInStyles22Su} adopts a shadow guidance which retrieves sketches from a database rather than using the generated images directly. 

\vspace{1,6mm}
\noindent
\textbf{Sketch-based applications}. 
As a high-level abstract representation, sketches can be used as conditional inputs to generative models. 
Sketch-based systems allow users to intuitively obtain results in various applications, such as image retrieval~\cite{lee2011shadowdraw, LiuSSLS17, YuLSXHL16} and image manipulation~\cite{DekelGKLF18,FaceShop18,ArtWF20,DPS20}, simulation control~\cite{hu2019sketch2vf}, block arrangement~\cite{sketch2domino}, and 3D modeling~\cite{Fukusato2020Meshing, igarashi2001suggestive, igarashi1997interactive}. 
As for applications such as iSketchNFill~\shortcite{iSketchNFill2019}, which consider the sketching process as input, the generation quality is still limited and cannot be applied to high-quality anime portrait generation. 
Although there have also been attempts to generate high-quality faces with sketches (e.g., DeepFacePencil~\shortcite{DeepFacePencil20}), they do not take into account the case of sparse input sketches at the beginning of drawing process. 
On the other hand, the vast majority of S2I studies~\cite{DeepFaceDrawingChenS0XF20,Yang21ControllableS2I} target the generation of real images, but how these methods can be applied to the abstract artistic style for art drawing assistance has not been explored, which is the topic in this paper. 


\section{Stroke-level Disentanglement}\label{sec:Disentanglement}
\begin{figure}
\centering
\includegraphics[width=0.98\linewidth]{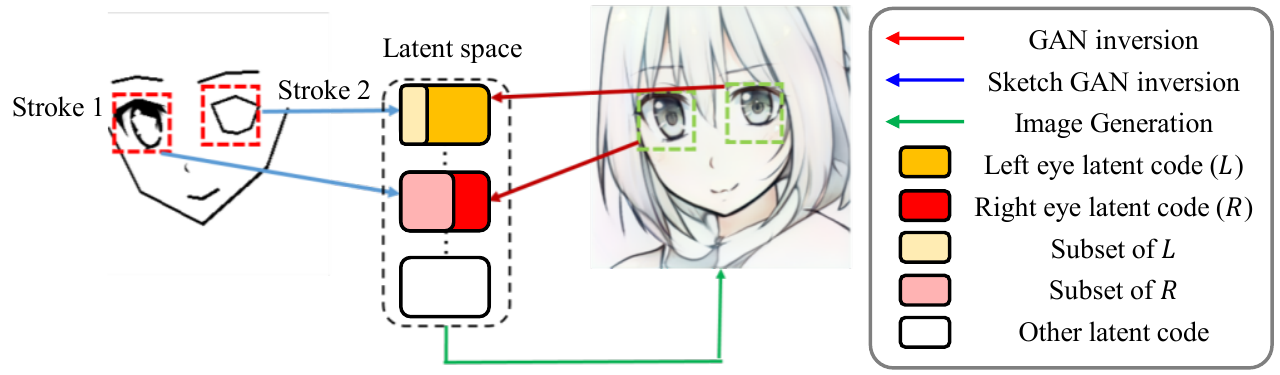}
\caption{Illustrating the stroke-level disentanglement.}
\label{fig:stroke-level-dis}
\end{figure}

The use of sketches to control shape properties of an anime face to achieve latent space exploration is called stroke-level disentanglement. 
We first explain its concept with a simple example (\autoref{fig:stroke-level-dis}). 
Given an image generated by StyleGAN with a fixed color latent code, the left/right eyes (green box) are mapped to $L$ and $R$ in the disentangled latent space with GAN inversion. Stroke-level disentanglement means that there is a sketch-GAN inversion coder for the rough sketch that allows Stroke~1 and 2 (red box) to be mapped to the subset of the corresponding latent codes $L$ and $R$, respectively. 
Note that the percentage of the latent code  (intersection of the same structural information $\div$ latent codes for a single facial part) of Stroke 1 to $L$ is higher than that of the latent code of Strokes 2 to $R$, because Stroke 1 contains more details. In addition, there may be a one-to-many relationship between the strokes and the latent code of different facial parts; for example, if a stroke contains shape information of both the left and right eyes at the same time, it will correspond to a subset of both $L$ and $R$ after encoding.

We formally describe the problem as follows. 
Let $\bm{P}$ and $\bm{S}$ denote the anime portrait domain and the sketch domain, respectively. $\bm{Q}$ is a subset of $\bm{P}$ that separates most representations of color information from structural information and can form a one-to-one mapping with S. Our sketch encoder learns a mapping $\bm{F: S \rightarrow Q}$ which can find the correct correspondence with increased drawing strokes. 
This mapping $\bm{F}$ is called ``sketch GAN inversion'' in this paper. 
The output during the drawing process should gradually converge and maintain high quality as the input strokes increase. Two main research questions need to be addressed: 
\begin{itemize}
\setlength{\leftskip}{-3mm}
    \item Q1. How does one learn a stroke-level disentangled mapping~$\bm{F}$ that allows strokes to locally match to the generated image?
    \item Q2. How can the aforementioned mapping not be affected by the stroke order?
\end{itemize}
Given a sketch consisting of a series of strokes $\{\bm{s_1,s_2,...s_n}\}$, 
 these two questions require that mapping $\bm{F}$ in ideal cases satisfies the following two conditions.
 
\vspace{1mm}
\noindent
\textbf{Stroke independence}. Assume that an image encoder that converts an anime portrait to completely disentangled structural latent codes $\{\bm{d_1,d_2,...d_n}\}$ corresponding to strokes one by one, then:
\begin{equation}
\begin{split}
\bm{F(s_i)}=\bm{d_i}
\end{split}
\end{equation}
where $i$ is the index of strokes ($i \leq n$). Note that each stroke can create a new partial sketch \{$s_i$\} that contains only one facial part. 

\vspace{1mm}
\noindent
\textbf{Stroke order invariance}. For any different stroke index $i,j \leq n$: 
\begin{equation}
\begin{split}
\bm{F(s_i|}&\bm{s_1,s_2,...s_{i-1},s_{i+1}...s_n)}\\
    &=\bm{F(s_j|s_1,s_2,...s_{j-1},s_{j+1}...s_n)}\\  &=\bm{F(s_1,s_2,...s_n)}
\end{split}
\end{equation}
where $\bm{s_i|s_1,s_2,...s_{i-1},s_{i+1}...s_n}$ means add stroke $\bm{s_i}$ to a sketch consisting of strokes $\{\bm{s_1,s_2,...s_{i-1},s_{i+1}...s_n}\}$. 
Note that we do not use any semantic label, and the inputs are monochrome sketches.

\begin{figure}
\centering
\includegraphics[width=0.98\linewidth]{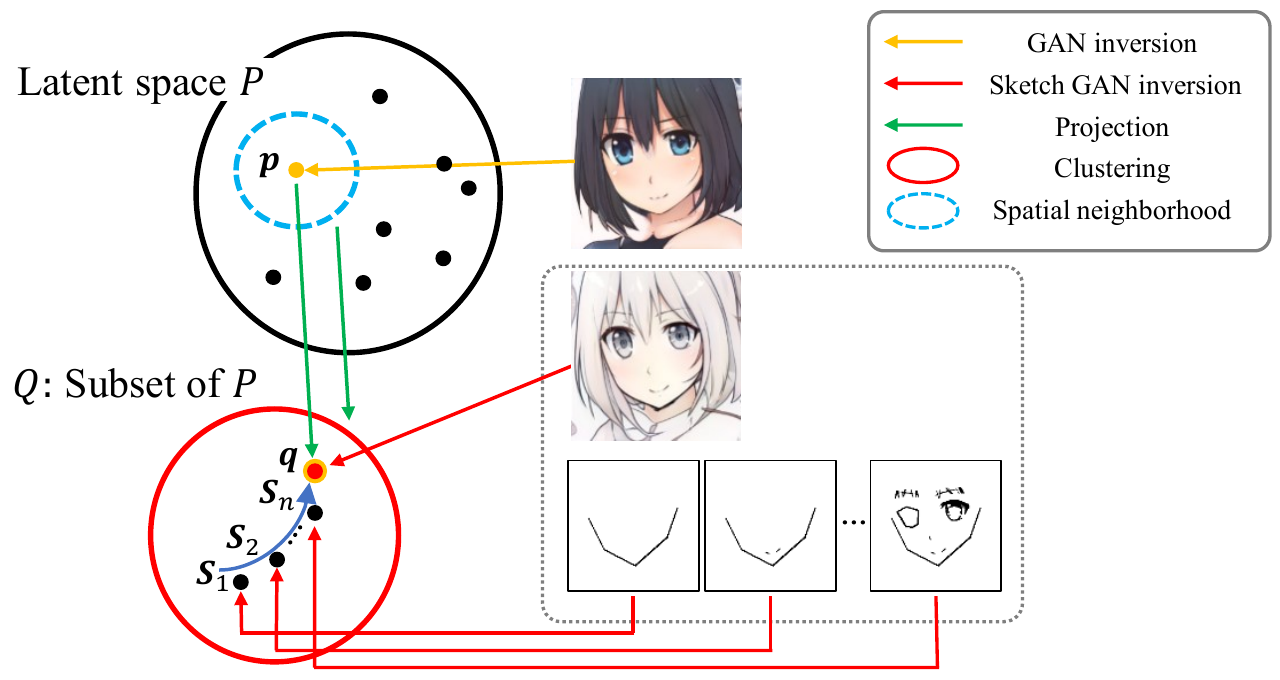} 
\caption{Illustrating the drawing process in the latent space.}
\label{fig:main_idea}
\end{figure}

\autoref{fig:main_idea} shows our core idea of simulating the drawing process and making the sketch with a higher degree of completion, closer to the original sketch, in the latent space limited to a neighboring region that can provide the answers to the above two questions. Given an image generated by StyleGAN, the point computed by GAN inversion in the latent space $P$ is $p$, and the point in the image fixed color latent code (first row in the gray dashed box) projected into the latent subspace $Q$ is $q$. Our drawing process simulation generates a sequence of simulated sketches (second row in the gray dashed box) from simple to complex, whose positions in $Q$ space are denoted as $S_1$ to $S_n$. The core idea is to learn a spatial neighborhood in $P$ whose projection in subspace $Q$ can make the sequence of points $S_1$ to $S_n$ gradually approximate the point $q$. 

\section{Proposed Framework}
\begin{figure}
\centering
\includegraphics[width=0.98\linewidth]{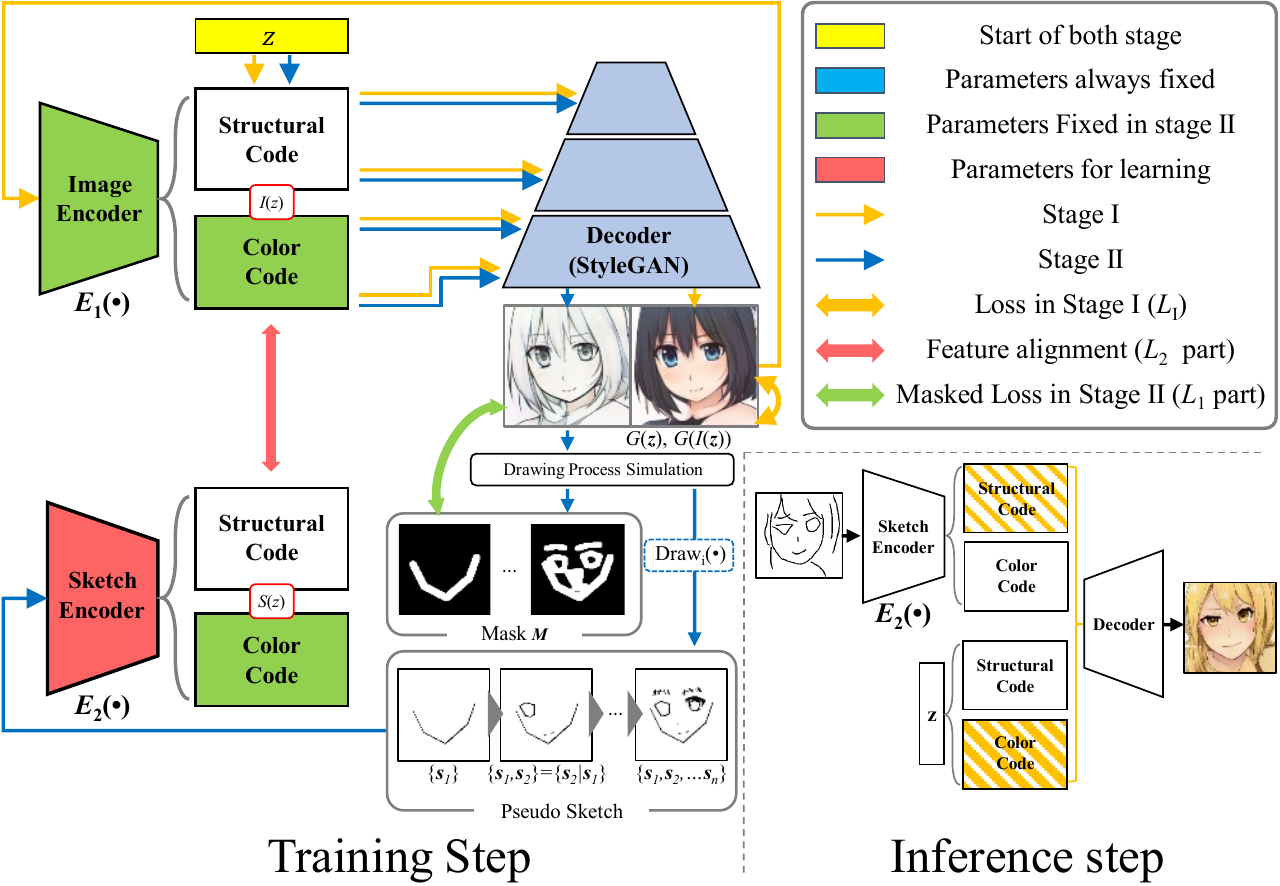}
\caption{Framework of the proposed method. 
The training step consists of two stages: (Stage I)~an image encoder for GAN inversion and (Stage II)~a sketch encoder for sketch GAN inversion where the image encoder works as a teacher.\vspace{-2mm}}
\label{fig:overview}
\end{figure}
An overview of the framework is shown in \autoref{fig:overview}. In the training step, we first trained an image encoder using the randomly-generated images from the decoder, which correctly projected the anime portraits back into the latent space (Stage I). 
Then, we rearranged the latent space vectors in this image encoder by simulating the drawing process, so that sketches with similar strokes retained more rational distribution when projected into $\bm{Q}$ (Stage II). 
In the inference step, we concatenated the structural codes derived from the sketch encoder with the color codes from the random Gaussian noise $z$, which is known as style-mixing. Note that once the decoder is determined, all data are derived from the randomly-generated images of that decoder, and no additional auxiliary database is required. Thus, this is an unsupervised learning  approach.

The training in Stage I is similar to previous work~\cite{psp21}. The difference is that we simply adopted the $L2$ loss between the original images from StyleGAN and the reconstructed images encoded by our image encoder.

\subsection{Drawing Process Simulation}
\begin{figure}
\centering
\includegraphics[width=\linewidth]{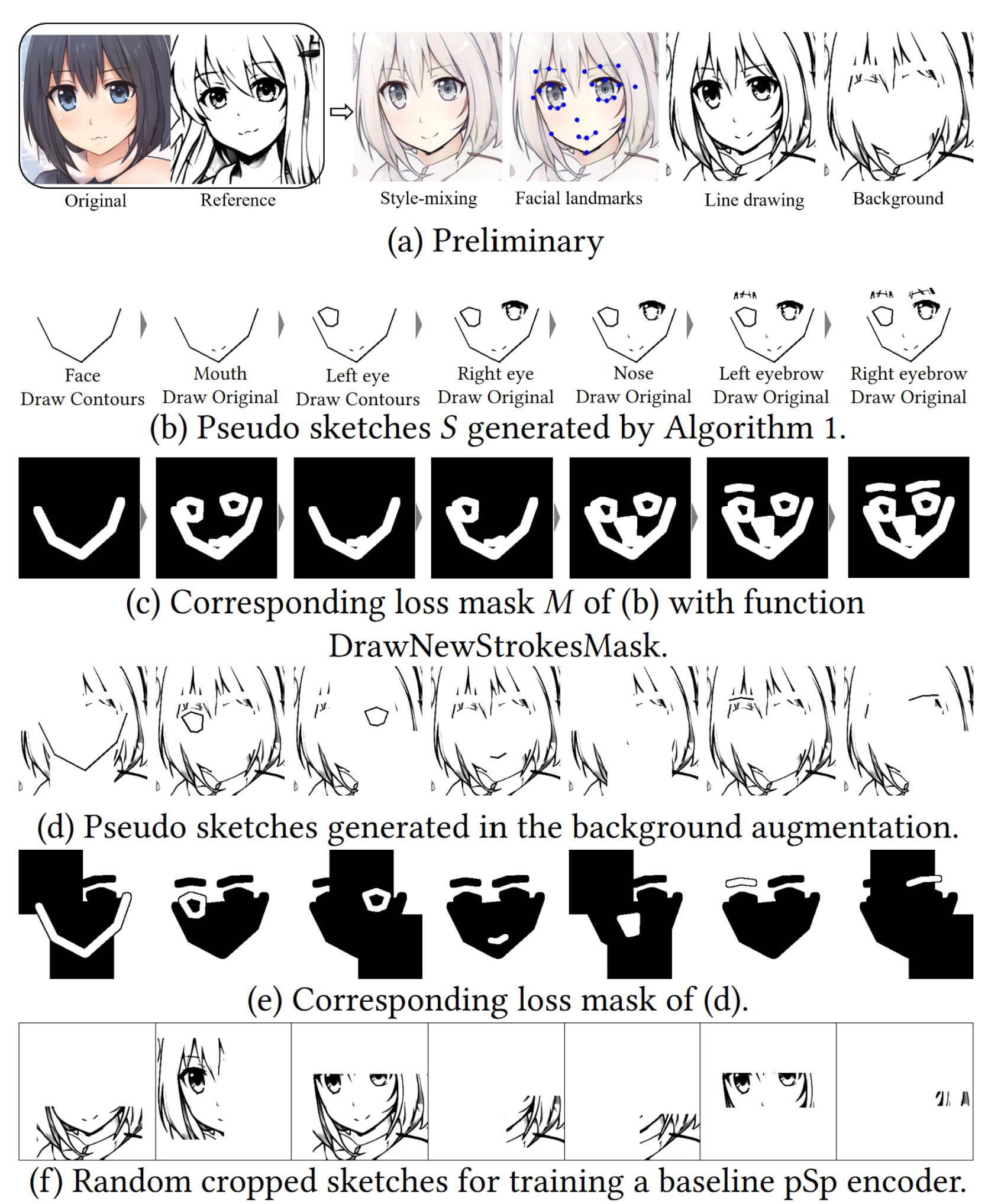}
\caption{An example of (b, c)~the drawing process simulation and (d, e)~background augmentation after (a)~a series of pre-processing for a randomly generated image (original) from StyleGAN. The corresponding choices of $RandomSelectOneStroke$ and $RandomDrawing$ functions for each stroke are shown in (b). 
In (c), $DrawNewStrokesMask(m_t, s)$ executes $DrawContours$ on the previous loss mask $m_t$ for each stroke $s$ to get a new loss mask, except for the nose (the nose area is determined by the landmarks in the center of eyes and the corners of mouth). For comparison, we also trained a baseline encoder with a random cropping strategy from the input line drawing in (f).}
\label{fig:drawingsmulation}
\end{figure}

In Stage II, the drawing process simulation automatically generates sketch-image pairs from StyleGAN. 
Before the drawing process simulation, we should first get a complete line drawing of the original anime portrait generated by StyleGAN as the simulation input. As shown in the first line of \autoref{fig:drawingsmulation}, we conducted a style-mixing between the original and the reference image so that most of the color information could be removed and obtained a complete line drawing from the style-mixing result with xDoG~\shortcite{XDoG11}. 
We then used a landmark detection technique for anime faces~\cite{anime-face-detector} to acquire information about the contours of each face part.
We simulated the intermediate results of the sketch process stroke by stroke using Algorithm~\ref{algo:simulation}. 
Facial landmarks are divided into $n$ types ($n=7$), and for each new stroke in $k$ iterations, a facial part is randomly selected as a part stroke $s \in C$ with RandomProcess and RandomDrawing as \autoref{fig:drawingsmulation}(b) shown. In the same iteration, the corresponding cumulative loss mask is drawn by the function $\operatorname{DrawNewStrokesMask}$~(\autoref{fig:drawingsmulation}(c)).
%
%

\vspace{1mm}
\noindent
\textbf{Background augmentation}. 
Since the hair and other parts could not be extracted by the anime face detection algorithm, we treated them as background. To increase stability, random cropping of the background image and random selection from facial contours were combined as augmentation data. The effects of this method are discussed in Section~\ref{sec:stability}.
At this point, we had a series of pseudo sketches for training in Stage II which is called ``feature alignment.''

\begin{algorithm}
	\caption{Drawing process simulation}
	\label{algo:simulation}
	\KwIn{Portrait image $P$}
	\KwOut{List of pseudo sketch $S$, List of loss mask $M$}
	Landmarks of portrait $ L \leftarrow \operatorname{FaceDectect}( P$)\; 
    Strokes of facial parts $C \leftarrow \operatorname{Resort}(L)$\;
    Number of $n \leftarrow len(C)$\;
    Temporary white image $p_t \leftarrow$ ones($P$.shape) $\times$ 255\;
    Temporary loss mask $m_t \leftarrow$ zeros($P$.shape)\;
		
	RandomProcess=[GaussianBlur(kernel size$=3\times3$); Dilate; Erode, KeepOriginal(None)]\;
	RandomDrawing=[DrawOriginal, DrawContours]\;
    $S \leftarrow \emptyset$ \;
    $M \leftarrow \emptyset$ \;
	\For{k=1:n}
	{
		Index $i=\operatorname{RandomSelectOneStroke}(C)$\;
		
		Part stroke $s$=$C$.pop($i$)\;


		$p_t= \operatorname{RandomDrawing}(\operatorname{RandomProcess}(p_t$ ,$ s $)) \;
  
		$S$.push($p_t$)\;
		$m_t$= $\operatorname{DrawNewStrokesMask}$($m_t$, $ s $) \;
		$M$.push($m_t$)\;
	
	}
	return $S$,$M$

\end{algorithm}

\subsection{Feature Alignment}
Given Gaussian noise $\bm{z}$, the input image of our encoder is $\bm{x}=G(\bm{z})$ and the output latent code $I(\bm{z})$, a special implementation of a point $p$ in $P$ (\autoref{fig:main_idea}), is then defined as: 
\begin{equation}
    I(\bm{z}):=E_1(G(\bm{z}))
\end{equation}
where $E_1$(·) and $G$(·) denote the image encoder and StyleGAN generator, respectively.
Then, 
our method for training an image encoder in Stage I followed the usual GAN inversion method.
The loss function $L_I$ we used in Stage I is as follows: 
\begin{equation}
    L_I=L_2(G(I(\bm{z})), G(\bm{z}))
\end{equation}
Just by calculating the $L2$ distance between the input image and the reconstructed image, the image encoder can already learn inverse mapping very well. 
Similarly, we defined the output latent code of our sketch encoder as follows:
\begin{equation}
    S(\bm{z}):=E_2(\operatorname{Draw}_i(G(\bm{z})))
\end{equation}
where $E_2$(·) and $\operatorname{Draw}_i$(·) denote our sketch encoder and our drawing process simulation as described in Algorithm 1, which can convert the image $\bm{x}$ to a series of intermediate sketches of the drawing process and select the $i$-th sketch from among them.

In each iteration of training in Stage II, we can generate sketches $\bm{S}$ and corresponding loss masks $\bm{M}$ after our drawing process simulation. Then, the loss function is:
\begin{equation}
    L_S=L_1(G(S(\bm{z}))*M,G(\bm{z})*M)+L_2(I(\bm{z}), S(\bm{z}))
\end{equation}
$L_2(I(\bm{z}), S(\bm{z}))$ ensures that the sketch with a higher degree of completion is closer to the projection of the original in the latent subspace, while $L_1(G(S(\bm{z}))*M,G(\bm{z})*M)$ with a loss mask ensures the local similarity between the originals and the generated results.

\section{User Interface}

\begin{figure}
\includegraphics[width=\linewidth]{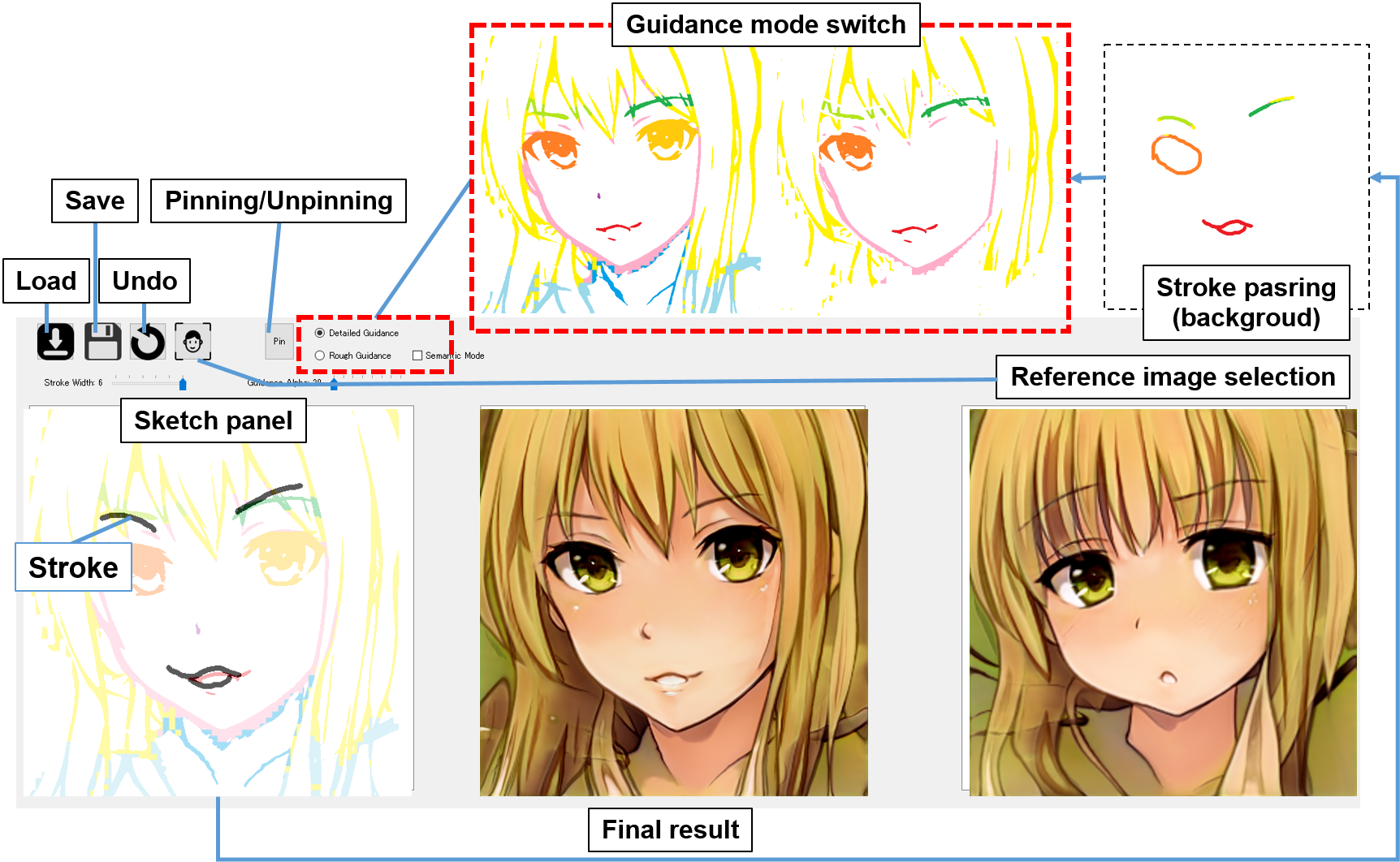}
\caption{User interface of the proposed AniFaceDrawing drawing assistance system. User guidance is generated using our sketch-based latent space exploration approach.}
\label{fig:c6UI}
\end{figure}

\autoref{fig:c6UI} shows our drawing assistance system. 
The system automatically records all the vertices of strokes and the stroke order, and converts the strokes into a raster image and corresponding guidance display on the sketch panel in real-time.
Similar to ShadowDraw~\shortcite{lee2011shadowdraw}, this system provides two types of guidance: ``rough guidance'' and ``detailed guidance,'' which users can switch at any time. 
Different colors denote semantics of the generated line drawing and we calculated the colorized guidance with a combination of few-shot semantic segmentation~\cite{PANet:Wang19} and one-shot learning for StyleGAN controlling~\cite{OneshotScribbles:Endo22} (Section 2 of the supplementary material contains more details). 
Detailed guidance shows the full-face portrait to the user as a prompt, while  rough guidance shows the user a part of the face that has been drawn roughly or will be drawn soon as a prompt by predicting the the user's drawing progress. In rough guidance mode, only a single semantic part (color) will be shown according to the moving mouse point. 
Both are useful and high-quality: detailed guidance allows the user to understand the overall layout of the face to draw, 
and rough guidance allows the user to focus on drawing the local parts of the face.
If the user is satisfied with the current guidance and does not want to change it any further for a sketch trace, 
he/she can press the ``Pin'' button to realize this purpose.  
When the sketch is completed, users can generate the final color image by clicking on the ``reference image selection'' button (face icon) to select the coloring style from the reference images.
In contrast, with the ``Eraser'' tool, users right-click on a stroke and the system erases it. In addition, the ``Undo'' tool can delete the last stroke from the stroke list.  
Note that our system can also load (or export) user-drawn strokes by clicking the ``Load'' (or ``Save'') buttons.

\section{Experiments and Results}

\subsection{Implementation Details}
The image encoder and sketch encoder (\autoref{fig:overview}) adopted the pSp architecture~\shortcite{psp21}. We chose layers 1-8 in  $\bm{\mathcal{W}}+$ space as the structural code and layers 9-18 as the color code. We adopted a Ranger optimizer and set the learning rate to 0.0001. 
As a training environment, NVIDIA RTX3090 GPU was used to train our encoders on the Linux platform. 
Then, a workstation with Intel Core i7 8700, 3.20 GHz, NVIDIA RTX1070 GPU, and 64GB RAM on the Windows 10 platform was used as the testing environment. The input and output image resolution was 256$\times$256 and 512$\times$512, respectively; 35,480 iterations with batch size 8 in Stage I and 3,808 iterations with batch size 16 (7+9 for  ~\autoref{fig:drawingsmulation}(b, d)) in Stage II. Because data were randomly generated by $z$, ``max epochs'' is replaced by  ``iterations.'' 

\subsection{Stability Testing}
\label{sec:stability}
To test the stability of our sketch encoder during the freehand sketching process and verify the stroke-level disentanglement conditions, we first performed the following experiments. 
\begin{figure}
\includegraphics[width=0.95\linewidth]{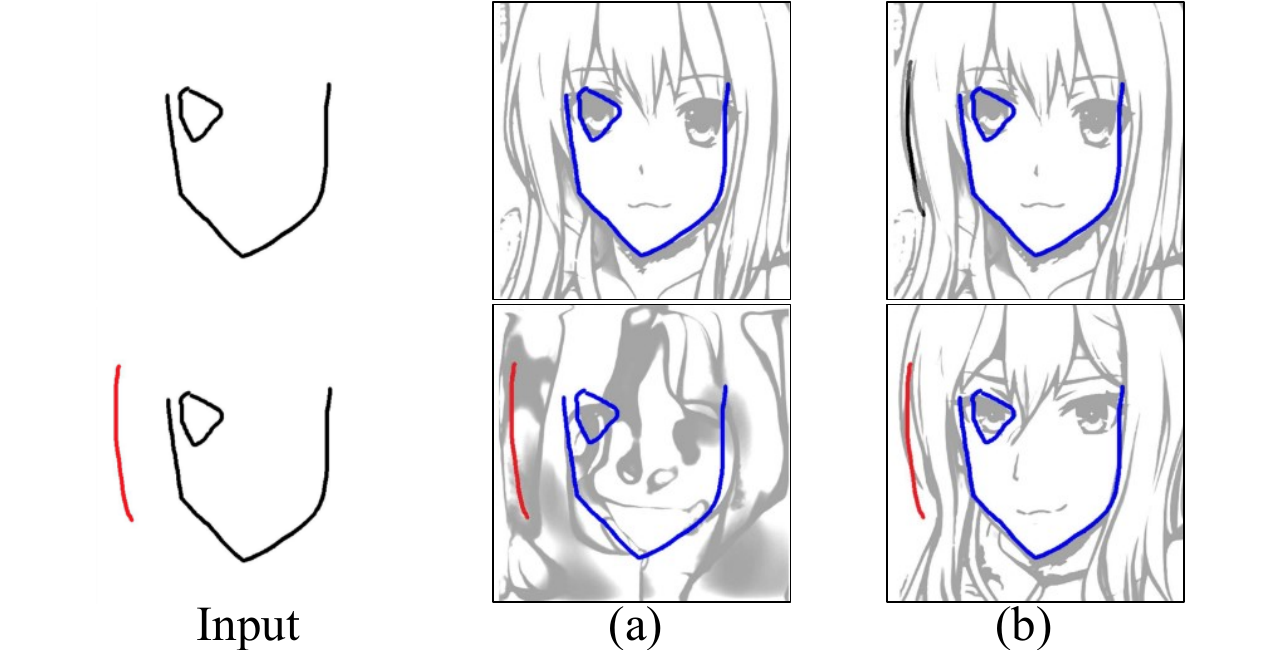}
\caption{Comparison of training (a)~without background augmentation and (b)~our training strategy in Stage II. 
When a red stroke is added, the result from the sketch encoder training without background augmentation is highly degraded.}
\label{fig:bg}
\end{figure}


\vspace{1mm}
\noindent
\textbf{Without background augmentation}. The effect of background augmentation is shown in \autoref{fig:bg}. If sketches in the drawing process (\autoref{fig:drawingsmulation}(b, c)) are trained without considering the background, the sketch encoder cannot correctly understand strokes associated with the background and project them near the correct position. 

\begin{figure}
\centering\includegraphics[width=0.95\linewidth]{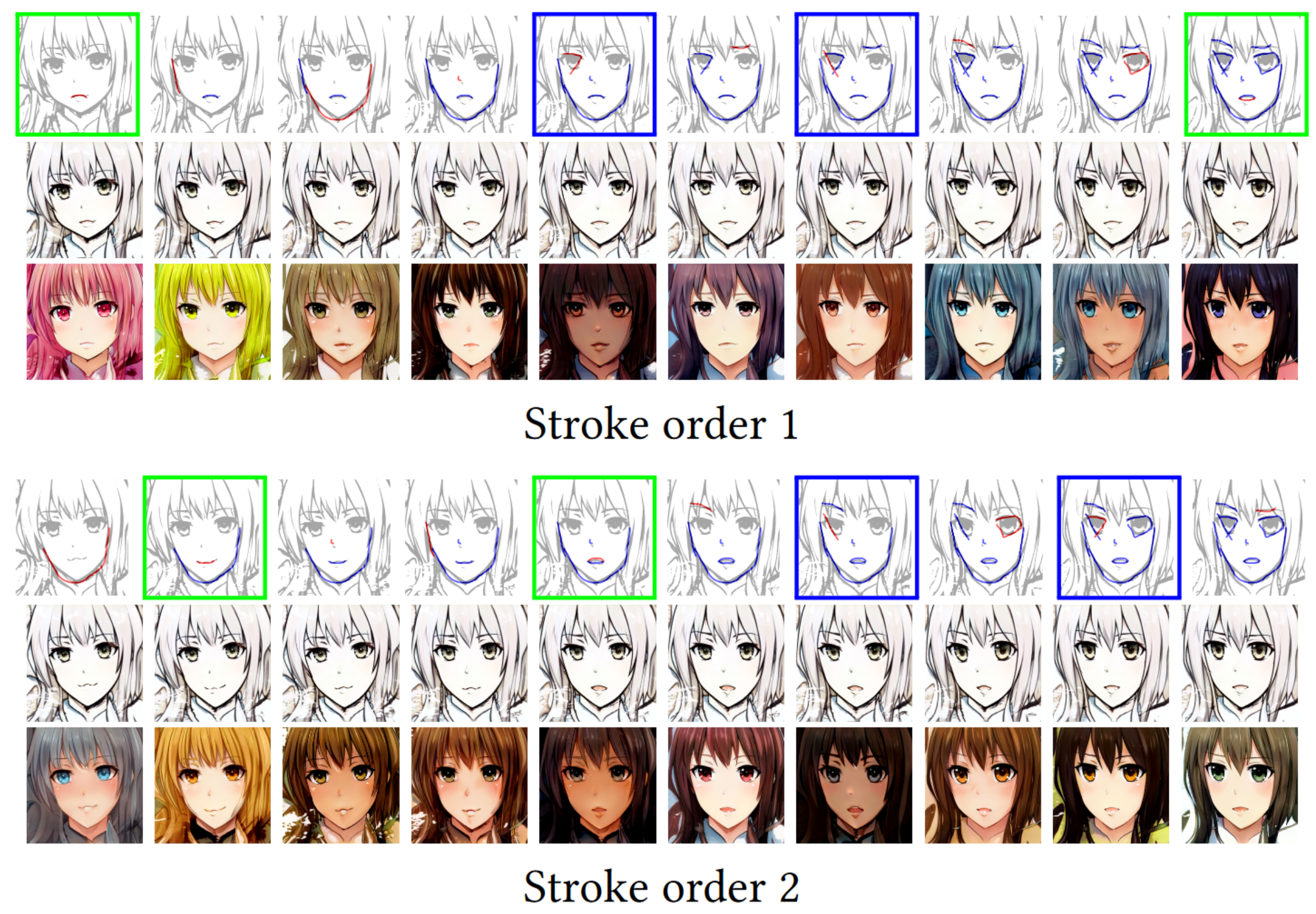}
\caption{The influence of stroke order and multiple strokes for one facial part (e.g., mouth (green) and left eye (blue)) in AniFaceDrawing.}
\label{fig:order}
\end{figure}

\vspace{1mm}
\noindent
\textbf{The influence of stroke order and multiple strokes on one facial part}. 
\autoref{fig:order} compares the intermediate process of the same sketch with different stroke orders.  It can be seen that the final results are not very different, but the intermediate processes maintain some diversity. This figure also shows that even if only one stroke is used for each part of the face during training, the generated guidance matches well when the user uses multiple strokes for the same part (e.g., the left eye and mouth).

\vspace{1mm}
\noindent
\textbf{``Bad'' stroke}. If only some parts of a stroke contain valid information, then the stroke is considered as ``bad.''  
In freehand sketching, ``bad'' strokes are not uncommon. The results generated by our method provide a suitable match to the valid part of such ``bad'' strokes. 
For example, the strokes representing the left eye in \autoref{fig:order} form a triangle, a shape that is not natural  for representing the eye contour, while the generated result is still reasonable. 
Another example is the first stroke, which partially matches the normal face contour (see \autoref{fig:c5-cmp1Quality}), but our approach still manages to capture this information and ignore the meaningless part of this stroke. 

\subsection{Qualitative Results}

\label{sec:quality_res}
 We found that there is no S2I synthesis technique for anime portraits due to the lack of paired data. 
 Therefore, we trained an additional sketch encoder for the complete sketch using a random cropping strategy with the state-of-the-art pSp encoder~\shortcite{psp21} as a baseline for a fair comparison. Some examples of input sketches in the training step for this baseline encoder are shown in \autoref{fig:drawingsmulation}(f). Except for this random cropping training strategy, which is used in iSketchNFill~\shortcite{iSketchNFill2019}, the hyperparameters and the architecture of the baseline network are the same as those in our sketch encoder. 
 From the comparison results (\autoref{fig:c5-cmp1Quality}), we verified that our method can provide consistently high-quality guidance that better matches the input during the sketching process. Meanwhile, when the input contains ``bad'' strokes, the generated guidance will provide a reasonable result rather than complete matching guidance. 

\renewcommand\wsize{0.32}

\subsection{Quantitative Results}
\begin{figure}
 \includegraphics[width=0.98\linewidth]{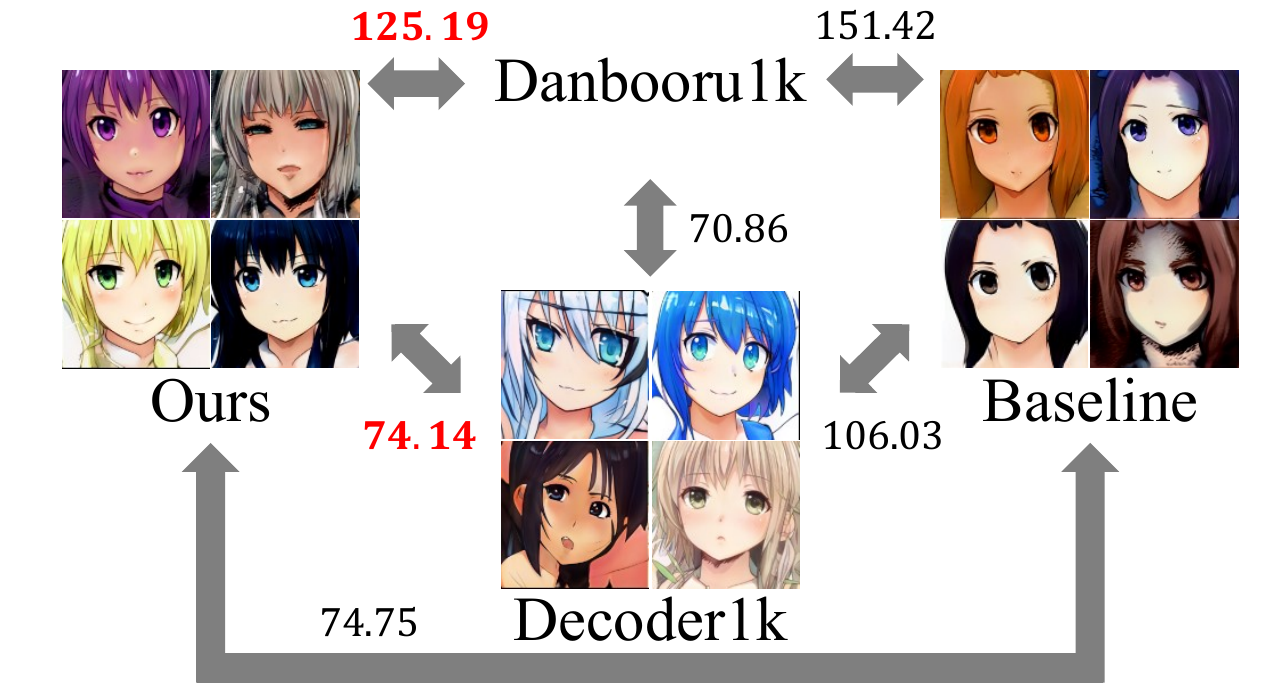}
\caption{Samples from the different datasets or approaches and the FIDs of each. }
\label{fig:c5fid}
\end{figure}

\begin{table}
\caption{FID scores of baseline and our approaches. As a reference, the FID between Decoder1k and Danbooru1k is 70.86.}
\centering
\begin{tabular}{c|c|c|c}

\toprule
DBName &  Ours & Decoder1k & Danbooru1k \\ 
\midrule
Baseline  & 74.75 & 106.03 & 151.42\\
Ours  & - & \textbf{74.14} & \textbf{125.19}\\
\bottomrule
\end{tabular}

\label{tab:c5FID}
\end{table}

Our approach was evaluated from two aspects: the quality of the generated images, and the match between input and output. 
The quality of image generation affects both the quality of the guidance received by the user and the evaluation of the final generated result, so it was necessary  to measure this indicator quantitatively in addition to the subjective evaluation of the user.
For similar reasons, the match between the input sketch and the guide needed to be measured quantitatively to ensure that the validity of our approach was subjectively and objectively consistent.
To evaluate usability and satisfaction, a user study was conducted for the overall system.

\begin{figure}[h]
\centering
\includegraphics[width=\linewidth]{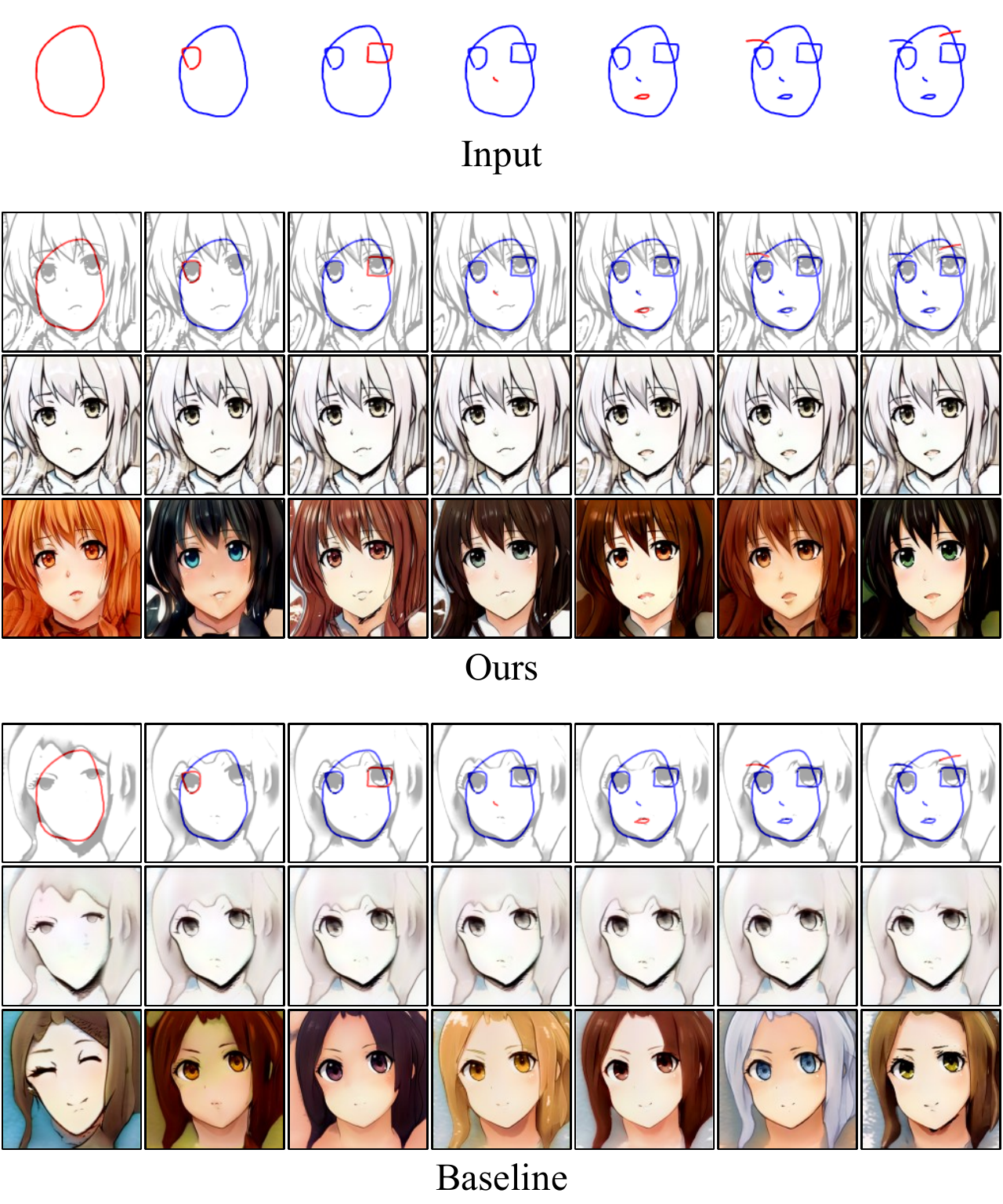}
\vspace{-3mm}
\caption{Qualitative comparison with the same input sketch sequences. A red stroke represents the last stroke in a sketch.}
\label{fig:c5-cmp1Quality}
\end{figure}

\vspace{1mm}
\noindent
\textbf{Quality of generated images}. Unlike normal S2I synthesis, this work was dedicated to the stability of matching rough sketches and intermediate results throughout the drawing process. 
To evaluate the degree of matching  between strokes and hints during the drawing process, this work used Fréchet Inception Distance (FID) to measure the gap between the generated images: 
First, users were asked to draw 10 sketches and record 177 images generated by our method as a database ``Ours,'' the results generated by the baseline method with the same input as a database ``Baseline,'' 1,000 randomly-generated images using StyleGAN in our decoder as a database ``Decoder1k,'' and 1,000 randomly-selected images from the Danbooru database~\cite{danbooru2019Portraits} as a database ``Danbooru1k.'' The FIDs between them are shown in Table~\ref{tab:c5FID}. It can be seen that, in line with the observation in the qualitative results, our method generated better-quality images 
-- similar to the images generated by the decoder in Decoder1k as well as to the real images in Danbooru1k. \autoref{fig:c5fid}  shows samples from each dataset or approach mentioned above, which makes results more intuitive.

\begin{table}
\centering
\caption{The average of different metrics from the proposed method and the baseline method (ours~/~baseline). 
At the beginning of the drawing process, the input sketches are usually sparser, which makes it difficult to generate matching results. Thus, the average recall score $r$ of the first $k$ strokes is more important.}
\label{tab:C6AP}
\resizebox{\linewidth}{!}{
\begin{tabular}{c|c|c|c|c|c}
\toprule
\diaghead{Metrics ww k}{Metrics}{$k$}& 1& 3& 6& 9 & whole process\\
\midrule
$p$  & 0.04/0.03& 0.04/0.05& 0.05/0.07& 0.07/0.08 & 0.12/0.12 \\
$r$   & \textbf{0.48}/0.40 & \textbf{0.46}/0.39& \textbf{0.45}/0.38& \textbf{0.43}/0.37 & \textbf{0.39}/0.31\\
$F1$  & 0.07/0.05& 0.07/0.09& 0.09/0.11& 0.11/0.13 & 0.17/0.16 \\
\bottomrule
\end{tabular}
}
\end{table}


\vspace{2mm}
\noindent
\textbf{Matching the generated image to the input sketch}. 
To evaluate the match between the input sketch and the generated guidance, 
sketch-guidance matching can be thought of as a prediction problem. 
Although neither sketch $S$ nor the generated line drawing $L$ is reliable enough as ground truth, the input sketches are regarded as ground truth here because what we are concerned about is how the system will cater to the input with guidance. 
The evaluation indicators precision $p$, recall $r$, and $F1$ score for sketch-guidance matching are defined in Section 1 of our supplementary material. 



\begin{figure}[t]
\centering
\includegraphics[width=1.0\linewidth]{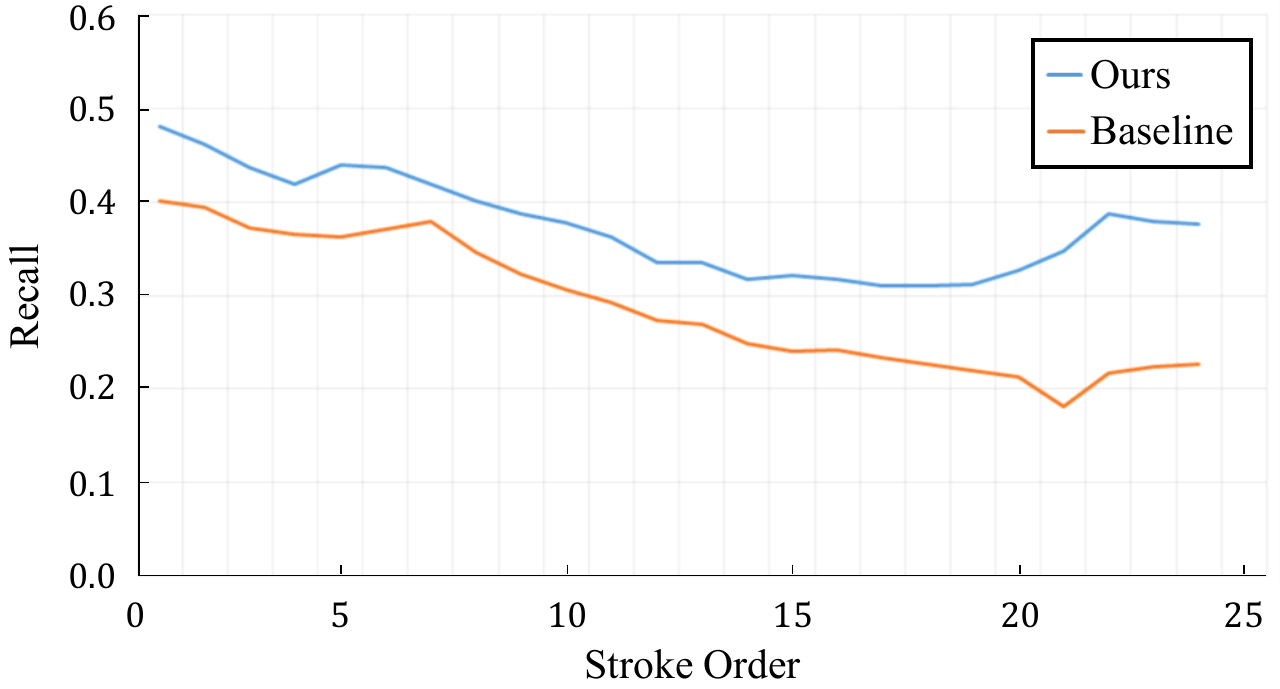}
\caption{Recall comparison as strokes increase. The average score of our method was higher than that of the baseline method.}
\label{fig:c6recall}
\end{figure}

\autoref{tab:C6AP} shows the comparison results between the proposed method and the baseline method (described in Section~\ref{sec:quality_res}). 
It is even more important to provide high-matching guidance in the early stages of drawing when strokes are sparse. 
Therefore, the average recall rate was calculated for the first 1, 3, 6 and 9 strokes and for the whole sketch process~($\infty$). 
We believe that the average recall is the best numerical description of sketch-guidance matching, as it is the proportion of overlapping areas of monochrome sketches and guidance over all current input sketches, and our results consistently outperformed and agreed with the baseline method in this metric throughout the drawing process, as shown in \autoref{fig:c6recall}.
This result was also consistent with the observation of the qualitative comparison (see \autoref{fig:c5-cmp1Quality}). However, slightly lower values of our method to the evaluation metrics $p$ and $F1$ compared to those of the baseline, indicate that our method provides more details in the guidance generation. 
This experiment demonstrated that the guidance generated by this system can better match the input sketch, both at the beginning and throughout the drawing process. 
 Based on the above results, recall can be considered a valid metric for measuring the match between the input sketch and output guidance. 

\vspace{-2mm}
\section{User Study}
\label{sec:c6_userstudy}
To verify the effectiveness of our anime-style drawing assistance system, we invited 15 participants (graduate students) to participate in user study. All participants were asked to draw anime-style portraits online using a remote mouse control. 
They were asked to draw the portraits freely and aimlessly, trying to draw as much detail as possible. 
Each user conducted anime portrait drawing twice: the first time to experience the whole process of drawing creation to familiarize themselves with the operation until they got used to this system and felt comfortable; and the second time, the participants completed the whole process independently. 
We instructed all participants on how to use AniFaceDrawing with user manuals (in the supplementary video). 
Before the hands-on experiments, they were asked to watch a tutorial video. 
All participants were required to carefully draw and select the most anticipated references for local guidance from several generated candidates after completing the global stage. 
If the generated guidance met their wishes and expectations, participants were required to press the ``Pin'' button to draw carefully to refine the input sketch. 
Participants could select a reference image for color portrait generation at any time during their drawing until they were satisfied with the results. 
Finally, they completed the questionnaire after finishing the second drawing. 
\subsection{Questionnaire Design}
\begin{table}
\caption{Custom questions in our user study.}
\label{table:c6allquestion}
\resizebox{\linewidth}{!}{
\begin{tabular}{l|l|c|c}
\toprule
\# & Question & Mean & SD   \\ 
\midrule
Q0 & How would you rate your drawing skills for anime/real faces ? & 2.07 & 1.10 \\ \hline
Q1 & Does the guidance match your sketch overall ?                       & 4.07 & 0.26 \\ \hline
Q2 & Does the guidance match your sketch when drawing the mouth ?        & 3.40 & 1.12 \\ \hline
Q3 & Does the guidance match your sketch when drawing the left eye ?     & 3.93 & 0.88 \\ \hline
Q4 & Does the guidance match your sketch when drawing the right eye ?    & 4.00 & 0.76 \\ \hline
Q5 & Does the guidance match your sketch when drawing the nose ?         & 4.07 & 0.80 \\ \hline
Q6 & Does the guidance match your sketch when drawing the hair ?        & 3.47 & 1.51 \\ \hline
Q7 & Does the guidance match your sketch when drawing the facial contours ? & 4.07 & 0.80 \\ \hline
Q8 & For your sketch and guidance, which facial balance is more reasonable ?      & 3.87 & 1.06 \\ \hline
Q9 & What is the quality of the guidance ? & 4.00 & 0.53 \\ \hline
Q10 & Does the guidance maintain high quality in your sketching process ?         & 4.27 & 0.46 \\ \hline
Q11 & Is the rough semantics guidance mode helpful for your drawing ?                        & 3.93 & 0.70 \\ \hline
Q12 & Is the detailed semantics guidance mode helpful for your drawing ? & 4.13 & 0.74 \\ \hline
Q13 & Are you satisfied with the final coloring results ? & 3.93 & 0.59 \\ \hline
Q14 & Does the guidance follow your will ? & 3.87 & 0.35 \\ \hline
Q15 & Are you satisfied with the final sketch result ? & 3.87 & 0.35 \\ 
\bottomrule
\end{tabular}
}
\end{table}
\begin{figure}[t]
\centering
\includegraphics[width=1.0\linewidth]{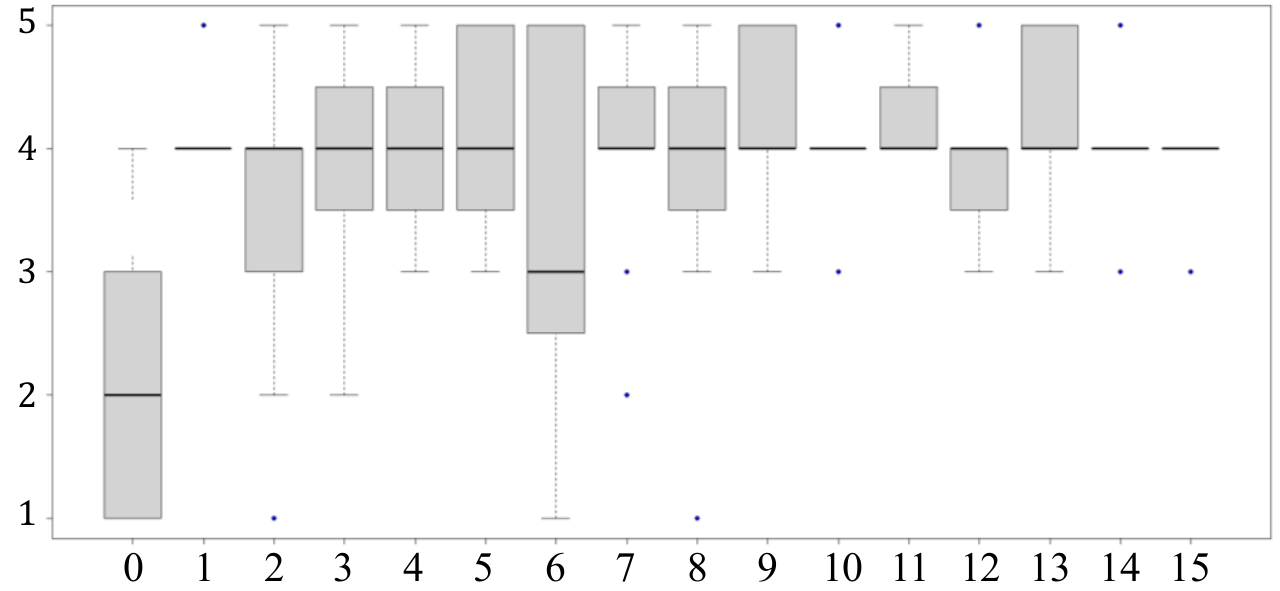}
\caption{Boxplots of custom questions in our user study. Questions Q0 to Q15 correspond to those in Table~\ref{table:c6allquestion}.}
\label{fig:c6_Plot}
\end{figure}
Our questionnaire consists of three parts:  the system usability scale (SUS)~\cite{Bangor08SUS}, creativity-support index (CSI)~\cite{Carroll09CSI}, and  a set of custom questions shown in Table~\ref{table:c6allquestion} to investigate the relationship between user satisfaction and guidance matching.
In the SUS, 10 questionnaire items were set up to capture subjective evaluations of the system usability. A five-point Likert scale was used in the evaluation experiment. 

Since the purpose of this work is to support user drawing creativity, the CSI is used to quantitatively evaluate the effectiveness of the proposed method. 
The CSI score defines the creativity of the tool with six factors: collaboration, enjoyment, exploration, expressiveness, immersion, ``results worth effort,'' and is scored with a maximum of 100 points. 
Here, the  ``Collaboration'' factor was set to 0 (not applicable) because there is no collaboration with another user in our task, as users completed the art drawing independently.

\newcommand\aniCSI{77.69}
\newcommand\realCSI{79.07}

\vspace{-2mm}
\subsection{Results}
\label{sec:c6_result}
This section discusses the visual results of AniFaceDrawing from users and user feedback from the user study. 
Table~\ref{table:c6allquestion} shows the mean and SD of each question in our customized questionnaire, while 
\autoref{fig:c6_Plot} shows the corresponding boxplots for these questions.


\vspace{1mm}
\noindent
\textbf{Visual results}. 
\begin{figure}[t]
\centering
\includegraphics[width=0.92\linewidth]{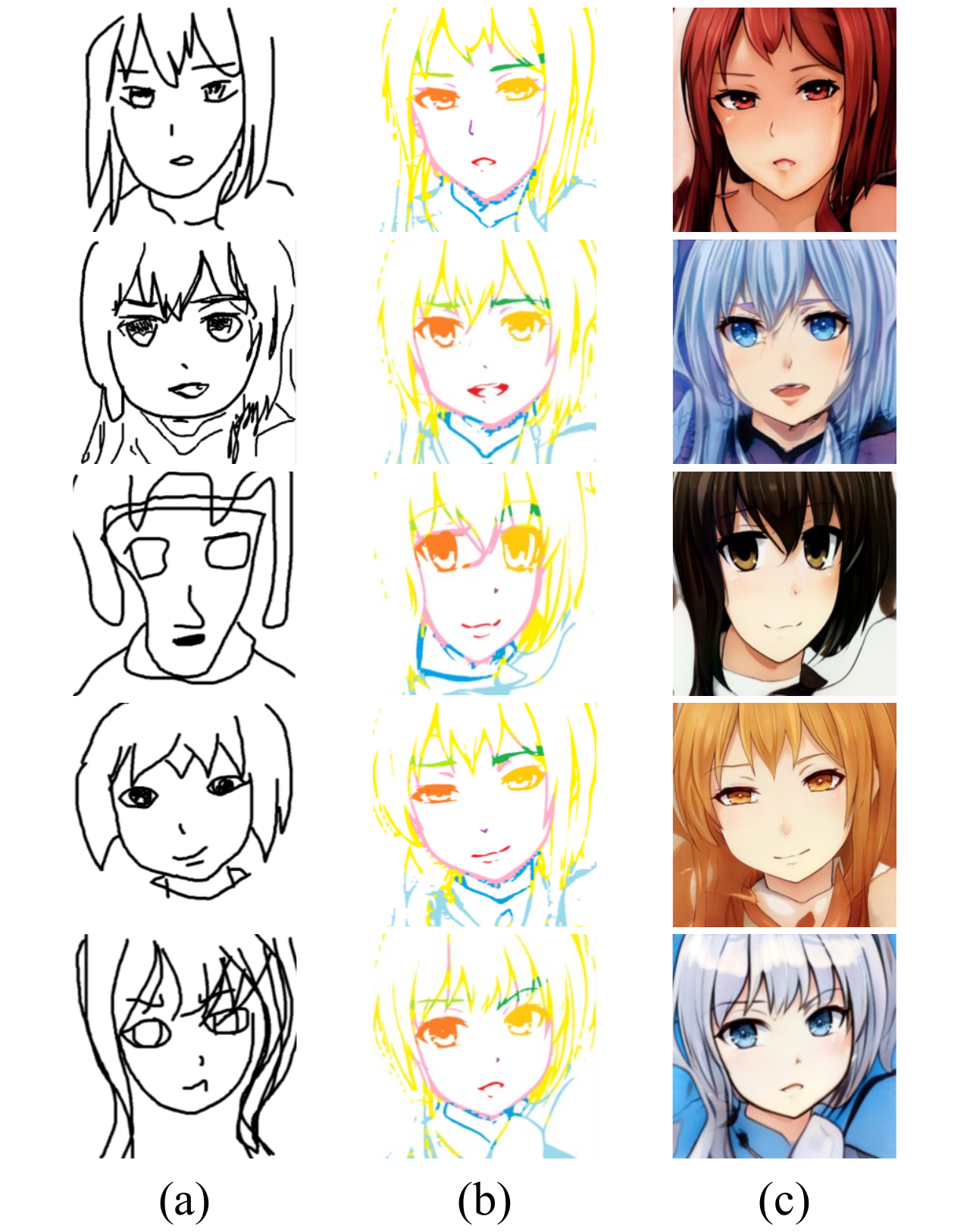}\\
\caption{Visual results from the user study. (a)~the final user sketches, (b)~the  guidance in detail mode, and (c)~the generated color drawings from (a) after the final reference image selection.}
\label{fig:c6_VisualResults}
\end{figure}

\autoref{fig:c6_VisualResults} shows some examples of the results generated in the user study. Our system could successfully transform the user's rough sketches into high-quality anime portraits. According to Q0 in our custom questions, 86.66\% of participants thought their drawing skills were not good enough (less than or equal to 3) for anime portrait drawing. 
As shown in  \autoref{fig:c6_VisualResults}, it can be concluded that even novices can make reasonable sketches with the help of the system and end up with high-quality color art drawings.

\vspace{1mm}
\noindent
\textbf{System usability}. The average score for SUS drawing assistance for anime style  was 73.84 ($SD=20.04$). The upper and lower limits were 90 and 65, respectively. 
In addition, participants stated that \textit{``Overall it's a good tool for those like me who do not have much drawing skills, and it's easy to use in terms of guidance generation and color selection''} and \textit{``I was not familiar with the operation when I first experimented, but I got an amazing generated result in the second experiment.''}
From these result, the usability of our drawing assistance system could be considered ``good'' for the anime style. 

\begin{table}[t]
\centering
\caption{CSI Questionnaire results in the user study.\vspace{-3mm}}
\label{table:c6csi}
\begin{tabular}{c|cc}
\toprule
Terms  & \multicolumn{1}{c}{Mean} & \multicolumn{1}{c}{SD}  \\ 
\midrule
Collaboration & -  & -  \\ 
Enjoyment     & 27.93 & 10.09 \\ 
Exploration   & 29.11 & 11.27 \\ 
Expressiveness& 23.50 & 6.30  \\
Immersion     & 13.46 & 10.86 \\
Results Worth Effort& 22.54 & 11.65 \\ 
\midrule
CSI Score     & \multicolumn{2}{c}{\aniCSI}         \\ 
\bottomrule
\end{tabular}
\vspace{-2mm}
\end{table}

\vspace{1mm}
\noindent
\textbf{Creative support capability}. As shown in Table~\ref{table:c6csi}, the average scores on CSI for anime style is \aniCSI.
Although there is still room for improvement in terms of immersion and expressiveness, the system can be used to create sketch-based art drawings.  

\vspace{1mm}
\noindent
\textbf{Time cost}. After the user draws a stroke, AniFaceDrawing provides an average response time of $1.65$ seconds for guidance generation. This response time appears to be too long for the user. Someone said, \textit{``One small problem is the not-so-short wait time after each stroke is completed.''} 
This also affected the immersion score in Table~\ref{table:c6csi}.
To improve the user experience, the calculation time needs to be further reduced.  
Although there was no time limit on our experiments, the average time to complete an experiment was about 9 minutes. 

\vspace{1mm}
\noindent
\textbf{User-perception match degree}. 
According to the results from Q1 to Q7 in Table~\ref{table:c6allquestion}, the average scores ranged from $3.40$ to $4.07$, which illustrates that our system can output relatively matching guidance to the input sketches during the drawing process.
In these questions, a score of 5 indicates a  \textit{``complete match''} and 1 point means a \textit{``complete mismatch.''} 
Although the statistics showed that the input and output matched relatively well, users disputed whether the hair in the input sketch and the generated image matched. 
The proponents commented, \textit{``The drawing assistance system performs better on hair and eyes, and can match well with the drawing person's draft to generate (anime portrait)."}  Critics said, \textit{``I tried to draw a double ponytail character, but couldn't achieve it''} and \textit{``Some special hairstyles cannot be generated by this system.''} 
The reason is that the stroke-level disentanglement is focused on facial contour features in the training step, and hair is trained with a random cropping strategy. 
Even so, most participants still tended to think that the sketch-hairs match is positive, which shows the generalization capability of our system at a certain level because we did not split the hair part into strokes to train in our training step. 

\vspace{1mm}
\noindent
\textbf{User-perception quality}. 
According to the results from Q8 to Q10 in Table~\ref{table:c6allquestion}, users  believed that the system consistently produced high-quality and reasonably balanced facial guidance throughout the drawing process in anime-style drawing assistance. 
This result is consistent with the results of the qualitative experiments in \autoref{fig:c5-cmp1Quality}, and they corroborate each other.

\vspace{1mm}
\noindent
\textbf{User satisfaction with guidance}. 
The results from Q11 to Q15 show that users generally agreed that our system provides good support for creating anime-style portraits, improving both the user’s own sketches and producing a desirable final color image according to their expectations.  
Considering Q1 to Q7, the consistency of these scores illustrates that our approach achieved the optimal match between sketch input and guidance output so that users are satisfied with our drawing assistance during the drawing process. 

\section{Conclusion}
We successfully re-ordered the feature vectors in latent space at the stroke-level by unsupervised learning with a drawing process simulation. The experiments demonstrated the stability and effectiveness of the proposed method. The experimental results show that our method can stably and consistently obtain high-quality generation results during freehand sketching, independent of stroke order and ``bad'' strokes. 
With our user study, AniFaceDrawing was proven to be effective and was able to create an anime portrait according to the users’ intentions. 
As a limitation, the matching of the input sketch for the hair part could be improved due to the training strategy. 
As the results generated by our method are completely dependent on the decoder—that is, the pre-trained StyleGAN—the decoder, in turn, restricts the types of images generated (refer to supplementary material for more details). 
For example, since our pre-trained model is trained on an anime portrait database selected from Danbooru, the generated results are all female. 
In addition, the current style is relatively constant: how to extract other styles from StyleGAN and make anime style more diverse and controllable will be explored in follow-up research. 
Meanwhile, how to expand the results with more styles, such as Ukiyo-e and painting, while keeping the strokes matching, is a promising topic for future work. 

\begin{acks}
This research was supported by the JAIST Research Fund, Kayamori Foundation of Informational Science Advancement, JSPS KAKENHI JP20K19845, and JP19K20316. 
\end{acks}


\bibliographystyle{ACM-Reference-Format}
\bibliography{ref.bib}
\balance
\end{document}